%% file: main.tex
\documentclass[10pt,twocolumn,letterpaper]{article}

\usepackage{iccv}
\usepackage{times}
\usepackage{epsfig}
\usepackage{graphicx}
\usepackage{amsmath}
\usepackage{amssymb}

\usepackage{enumitem}
\usepackage[parfill]{parskip}
\usepackage{gensymb}
\DeclareMathOperator*{\argmin}{arg\,min}
\usepackage{array}
\newcolumntype{P}[1]{>{\centering\arraybackslash}p{#1}}
\renewcommand{\deg}{$^{\circ}$}

\usepackage{siunitx}

\newcommand{\bff}{\boldsymbol{f}}

\newcommand{\br}{\boldsymbol{r}}

\newcommand{\bx}{\boldsymbol{x}}

\newcommand{\by}{\boldsymbol{y}}

\newcommand{\bPhi}{\boldsymbol{\Phi}}

\usepackage[pagebackref=true,breaklinks=true,letterpaper=true,colorlinks,bookmarks=false]{hyperref}

\iccvfinalcopy 

\begin{document}

\title{Angle Sensitive Pixels for Lensless Imaging on Spherical Sensors}
\author{Yi Hua\\
	Carnegie Mellon University\\
	{\tt\small yhua1@alumni.cmu.edu}
\and
Yongyi Zhao\\
Rice University\\
{\tt\small yongyi@rice.edu}
\and
Aswin C Sankaranarayanan\\
	Carnegie Mellon University\\
{\tt\small saswin@andrew.cmu.edu}
}


\maketitle

\begin{abstract}
We propose OrbCam, a lensless architecture for imaging with spherical sensors.
Prior work in lensless imager techniques have focused largely on using planar sensors; for such designs, it is important to use a modulation element, e.g. amplitude or phase masks, to construct a invertible imaging system.
In contrast, we show that the diversity of pixel orientations on a curved surface is sufficient to  improve the conditioning of the mapping between the scene and the sensor.
Hence, when imaging on a spherical sensor, all pixels can have the same angular response function such that the lensless imager is comprised of  pixels that are identical to each other and differ only in their orientations. 
We provide the computational tools for the design of the angular response of the pixels in a spherical sensor that leads to well-conditioned and noise-robust measurements.
We validate our design in both simulation and a lab prototype.
The implications of our design is that the lensless imaging can be enabled easily for curved and flexible surfaces thereby opening up a new set of application domains.
\end{abstract}

\section{Introduction}
\input{intro.tex}

\section{Related Work}
\input{prior.tex}

\section{A Lensless Spherical Imager}
\input{orbcam.tex}

\section{Experiments on Real Data}
\input{analysis.tex}

\section{Discussions}
\input{discuss.tex}

{\small
\bibliographystyle{ieee}
\bibliography{orbcam_bib}
}

\end{document}

%% file: intro.tex
Can we build a imager that is thin and conforming to a curved surface?
Such a imager would be invaluable for many applications. 
For example, it can be wrapped on a ball to produce a panorama. 
It can enable flexible robots to see their environment and can be pressed against human skin to accurately sense blood flows.
While recent advancements in flexible electronics \cite{nathan2012flexible} allow us to measure light intensity on a curved or flexible surface, it is very challenging to design lenses that are of a thin form factor and yet focus images on curved surfaces.
On the other hand, lensless imaging have delivered imaging solutions that are lightweight, compact and of thin form factor \cite{boominathan2016lensless}.
So far, lensless imaging techniques have been only developed for planar sensors, and we aim to incorporate flexible and curved sensors into lensless imagers.
In this paper, we take a step towards thin, surface conforming imager design, by proposing and analyzing the performance of a thin, lensless imager on a well-studied curved surface --- the sphere.

The missing piece of designing thin imagers that sense on a curved surface lies in producing the modulation element.
In planar imagers, a modulation element, traditionally a lens, is used to provide a diverse set of measurements so that the inverse problem of image recovery is well conditioned.
Consider the situation where a point light source is placed far from the sensor.
In absence of the modulation element, all pixels on a small planar sensor would all measure nearly identical measurements, as they receive nearly identical amount of light.
A thin modulation elements such as an amplitude masks \cite{Asif:17}, phase masks \cite{Antipa:18}, refractive \cite{tanida2001thin} and diffractive \cite{Gill:13} elements introduced diversity in the measurements so that the effective pixel response is rich and diverse enabling a well-conditioned inverse problem.
However, such elements are difficult to manufacture precisely for use on a curved or potentially flexible surface.

\begin{figure}[!ttt]
\centering
\includegraphics[trim={0.5cm 0cm 0cm 0cm}, clip=true, width=0.45\textwidth]{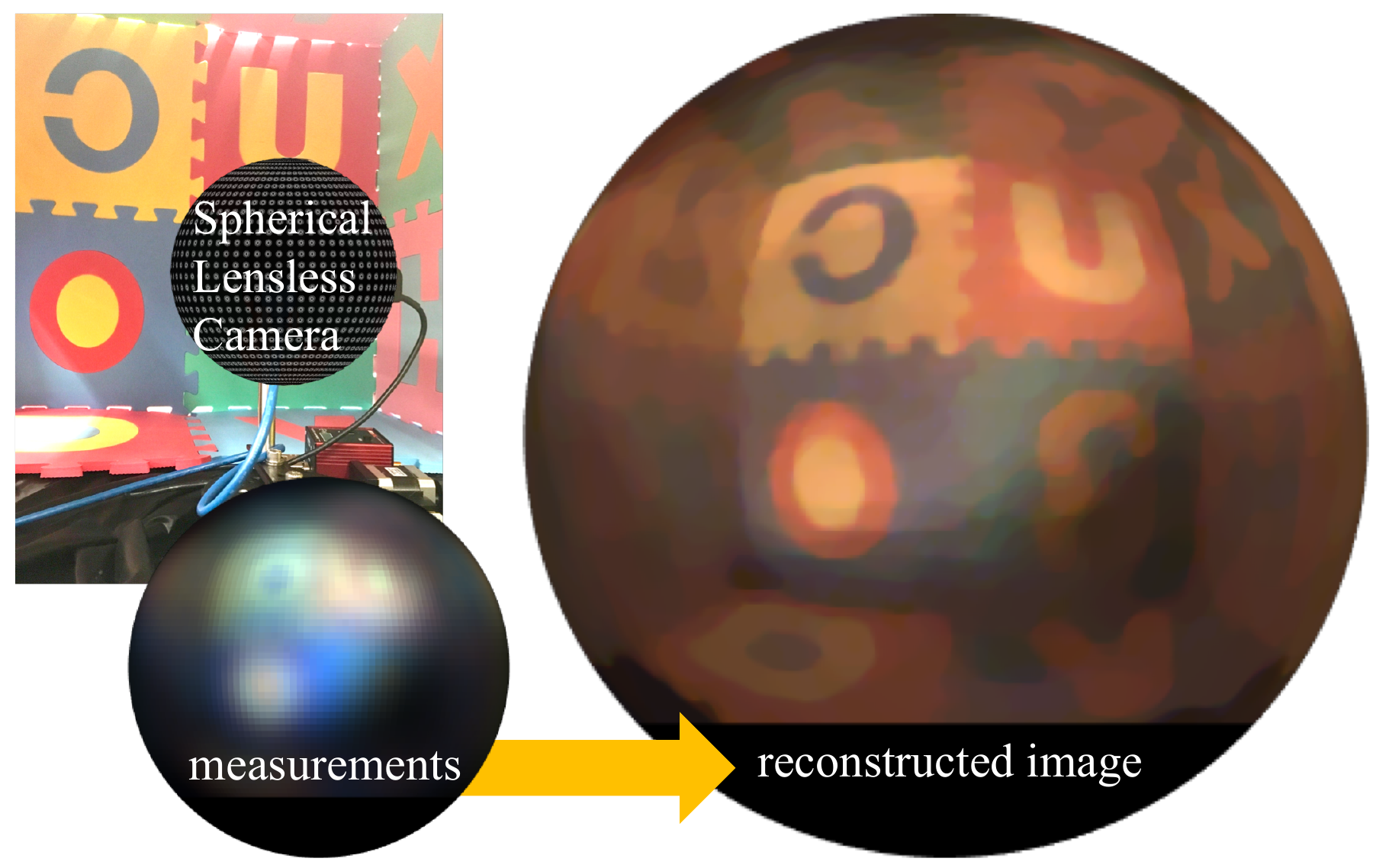}
\caption{We present a thin form-factor lensless imager on spherical surface. Imaging on a non-planar surface has many advantages, for example,  imaging  a large angle of view without radial distortion and vignetting. This image from inside of a cube built with foam mat captures $180^\circ \times 133^\circ$ angle of view with 6030 pixels.}
\label{fig:teaser}
\end{figure}

While designing a modulation element to induce diversity in measurements of a curved sensor is challenging, the curved surfaces present us with another source of measurement diversity that is absent in planar imagers, namely, \textit{the orientation of the pixels}.
On a curved sensor, the inherent diversity of pixel orientations is often sufficient to resolve the scene at a high angular resolution.
For example, if we had a spherical sensor comprised of photodiodes with very narrow cone of view, we could image a scene without any additional modulation.
But clearly, such a solution will have low light efficiency since we restrict the amount of light that enters each photodiode; this is especially true when we seek to resolve the scene at high resolution.

In this paper, we propose a design for lensless imaging using spherical sensors, where by tailoring the angular response of the pixels, we resolve the scene at high resolution without a commensurate loss in light throughput.
We envision a lensless imaging system with a bare spherical sensor, where each photodiode has identical but carefully designed angular response.
Unlike previous lensless cameras with planar sensors, we are able to have the same angular response at all pixels since spherical sensors have a  diversity of pixel orientations. 
As with previous lensless designs, the image of the scene is computationally recovered with or without the image priors from the measurements.

\begin{figure}
	\includegraphics[width=.45\textwidth]{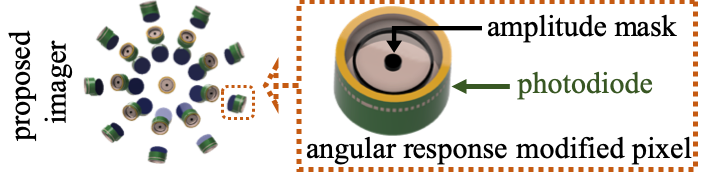}
	\caption{The proposed imager consists of a set of pixels with engineered angular response on a sphere.
This figure shows a design of using amplitude mask for modifying the angular response of pixel. }
	\label{fig:design}
\end{figure}

The proposed design where all pixels are engineered with the same angular response have many advantages against previous lensless imagers.
The angular-response engineering can be done during the CMOS process of the pixels.
The calibration procedure is simple, as we only need to measure the angular response of one pixel. 
Finally, regardless of the orientation of the pixels, the measurements always sample from the convolution between scene and angular response; this means the imager can be designed to be flexible, but the conditioning of the forward process is stable.

\paragraph*{Contributions.}
This paper presents a new design for lensless imaging on spherical sensor and make the following contributions:
\begin{itemize}[leftmargin=*]
\item \textit{Introduce pixel orientations as a source of measurement diversity.} As mentioned above, we exploit the insight that pixel orientations is in essence a form of modulation and provides diversity in measurements. This facilitates a simple design where all pixels on the spherical sensor can have the same angular response.

\item \textit{Design of optimal azimuth-symmetric angular response function.}
Under the assumptions of identical angular responses, we show that the sensor measurements are samples from isotrophic spherical convolution between scene and angular response, provided (a) the scene is sufficiently far away from the sensor, and (b) the angular response is also azimuthally-symmetric (with the azimuth angle calculated with respect to the pixel orientation).
There are many benefits to be derived from the choice of azimuthal symmetry.
First, the image formation can be computed inexpensively by scaling spherical harmonic coefficients of the scene.
Second, modeling as a spherical convolution provides analytical expressions for reconstruction error for known angular response.
As a consequence, the search for the optimal azimuthally-symmetric masks is tractable.
\item \textit{Validation via prototype experiments.} We verify the design of lensless  camera, and  in particular, the angular response function using a lab prototype. The prototype consists of a photodiode that is rotated around to simulate a spherical surface; the angular sensitivity of the photodiode is manipulated using an amplitude mask. Fig.~\ref{fig:teaser} shows a reconstruction from our prototype.
\end{itemize}
In addition to reconstructions from real data, we also provide numerous simulation results that highlight the properties of the proposed lensless camera.

%% file: prior.tex
We  discuss prior art in lensless imaging, as well as imaging on curved and spherical surfaces.

\subsection{Lensless Imaging on Planar Sensors}
Lensless imaging techniques have been proven effective in miniaturizing cameras.
Lensless imagers use a spatial light modulator as an alternative to traditional camera lens, and reconstruct captured scenes computationally. 
Most prior work model the sensor measurements, $\boldsymbol{z}$, as a linear combination of scene intensities, $\bx$, 
\begin{equation}
\boldsymbol{z} = \Phi \bx 
\label{eq:lensless}
\end{equation}
where $\Phi$ is the (linear) measurement operator.
The scene $\bx$ is recovered by inverting \eqref{eq:lensless}.

A common approach for implementing a well-conditioned measurement matrix is to cover the sensor with an amplitude or phase mask.
Asif \etal\cite{Asif:17} built a imager only 500~{\textmu m} thicker than the sensor by using an amplitude mask that is separable, so that the forward model can be simplified into two 1D convolutions.
Boominathan \etal\cite{boominathan2020phlatcam} and Antipa \etal\cite{Antipa:18}  employed thin transparent phase masks, and recovered 3D voxels in the scene by modeling measurement matrix as a sum of 2D cropped convolution.
Stork and Gill \cite{Gill:13} created ultra-miniature imagers ($\sim$100~{\textmu m}) with phase anti-symmetric spiral gratings integrated on planar sensors.
All of these designs rely on imaging with a planar sensor; we show that sensing on a curved surface allows the imager to be built with simple masks that are insufficient for imaging with a planar sensor.

The idea of engineering the angular response of sensors has some prior work in lensless imaging.
Wang \etal\cite{wang2009angle} manufactured an $8 \times 8$ angle-sensitive pixel   (ASP)  array on a CMOS sensor; here, each pixel in the array has a Gabor-like response function.
The measurements captured with the ASP sensor array were also used in combination with a lens to produce either high resolution 2D image or a low resolution 4D light field \cite{hirsch2014switchable}.
We further extend this idea of  angle-sensing pixels for imaging on non-planar sensors.

\subsection{Imaging on Flexible / Curved Surfaces} 
The last decade has seen many advances in flexible electronics and this has inspired the design of curved and flexible imaging sensors.
Previous solutions for focusing image onto curved surfaces can be split into three categories: with traditional monocentric spherical lens, with other types of refractive optical elements, and other lensless approaches.
An early work of curved sensor \cite{ko2008hemispherical} built an array of $16 \times 16$ photodiodes on a hemisphere, and more recently Guenter \etal \cite{guenter2017highly} have created mass-manufacturable high-resolution curved sensors by deforming 18 megapixel CMOS sensors.
Both results demonstrate curved sensor surfaces can improve image quality with traditional monocentric lens designs. 

Krishnan and Nayar \cite{krishnan2009towards} propose and analyze the design of a spherical imager consisting of a spherical sensor wrapped around a  ball lens.
Ball lenses have been used extensively with spherical sensors; however, such lenses add significantly to the bulk and weight of the imaging system and, further, cannot be used when the sensor is on the exterior of the sphere, i.e., the pixel are facing outwards.

Microlens arrays reduce the length of optical axis and lie closer to the curved sensor surface, and have been demonstrated useful for producing wide field-of-view images \cite{song2013digital}. 
Sims \etal \cite{Sims:16}\cite{sims2018stretchcam} designed elastic microlens arrays that can adapt with the curvature of the sensor. 
Alternatively, gradient refractive index lens designs were proposed for focusing a scene of a wide angle of view on curved sensor \cite{hiura2009krill}. 
They construct a prototype which can identify the angular position of a light source; however, the prototype was not extended to high resolution imaging. 

Compared to lens-based designs, lensless cameras place modulators even closer to the sensor surface, resulting in a more compact form factor.
Koppelhuber and Bimber \cite{Koppelhuber:17} provide a flexible imager design using flexible S\"{o}ller collimator and luminescent concentrators to guide light onto line-scan sensors.
However this imager assumes a focused image is formed on the luminescent surface which  is generally hard to obtain.
\subsection{Modeling Functions on the Sphere}

Understanding convolution on the sphere helps to clarify our choice of angular response function and placement of pixel orientations.
The underlying modeling and analysis is greatly simplified via the use of spherical harmonics. 
In addition to imaging applications, functions on the sphere have also been used to model light transport in rendering applications \cite{ramamoor2001, sloan2002}.
We give a brief description of the  properties most relevant to our results
while referring the reader to \cite{groemer1996geometric} for a detailed description.

\paragraph{Spherical harmonics}%
are a orthonormal basis for square-integrable functions $\mathcal{L}^2(\mathbb{S}^2)$ on the sphere. 
The spherical harmonics of degree $l$ order $m$ is a complex-valued function on the sphere, $l = 0, 1, \dots, \infty$ and $m = -l, \dots, l$,
\begin{equation}
Y_{l,m}(\theta, \phi) = \sqrt{\frac{2l+1}{4\pi}\frac{(l-m)!}{(l+m)!}} P_l^m(\cos \theta)e^{i m \phi},
\label{eq:ylm}
\end{equation}
where $P_l^m(x)$ are associated Legendre functions, $\theta \in [0,\pi]$, $\phi \in [0,2\pi)$ defines point on a sphere by its azimuth and altitude.
Any square-integrable function on the sphere $f\in \mathcal{L}^2(\mathbb{S}^2)$ can be composed into a sum of spherical harmonic basis weighted by $f_{l,m}$, its spherical harmonic coefficients
\begin{equation}
f(\textbf{r}) = \sum_{l=0}^{\infty} \sum_{m=-l}^l f_{l,m} Y_{l,m}(\textbf{r}).
\label{eq:f}
\end{equation}

\paragraph{Spherical harmonic coefficients} %
$f_{l,m}$ of a function $f(\boldsymbol{r})$ can be obtained by spherical harmonics transform,
\begin{equation}
f_{l,m} = \int_{\boldsymbol{r}\in \mathbb{S}^2} f(\boldsymbol{r}) Y_{l,m}(\boldsymbol{r}) d\boldsymbol{r}.
\label{eq:flm}
\end{equation}

\paragraph{Bandlimit on the sphere.} %
We say the function $f$ has bandlimit $L$ if its spherical harmonics coefficients $f_{l,m}=0$ for $l \geq L$. A function with bandlimit $L$ satisfies 
\begin{equation}
f(\boldsymbol{r}) = \sum_{l=0}^{L-1} \sum_{m=-l}^l f_{l,m} Y_{l,m}(\boldsymbol{r}).
\label{eq:f_bandlimit}
\end{equation}

\paragraph{Azimuthally-symmetric functions.}  
When a function $g \in \mathcal{L}^2(\mathbb{S}^2)$ is azimuthally-symmetric, i.e. $g(\theta, \phi) = g(\theta)$, it only contains spherical harmonics of order $m=0$. A azimuth-symmetric function $g$ satisfies 
\begin{equation}
g(\boldsymbol{r}) = \sum_{l=0}^{\infty}g_{l,0} Y_{l,0}(\boldsymbol{r}).
\label{eq:g_az}
\end{equation}

%
\paragraph{Isotropic convolutions on the sphere.} %
Convolving any signal with a bandlimited azimuthally-symmetric signal can be conveniently computed by scaling the signal's spherical harmonics coefficients. 
The associated deconvolution also simplifies to scaling spherical harmonics coefficients.
Formally, let $f(\boldsymbol{r})$ be a square-integrable signal on the unit sphere, and $g(\boldsymbol{r})$ be a square-integrable azimuthally-symmetric signal on the unit sphere with bandlimit of $L$, where $\boldsymbol{r} = (\theta, \phi)$ are points on the sphere with altitude  $\theta \in [0, \pi]$ and azimuth  $\phi \in [0, 2\pi)$. Their spherical convolution is
\begin{equation}
(f*g)(\boldsymbol{r}) = 
\int_{\boldsymbol{s}\in \mathbb{S}^2} g^*(\boldsymbol{r} \cdot \boldsymbol{s}) f(\boldsymbol{s}) d\boldsymbol{s},
\end{equation}
where $g^*(\boldsymbol{r})$ denotes conjugate of $g(\boldsymbol{r})$, and $\boldsymbol{r} \cdot \boldsymbol{s}$ is the inner-product between the vectors $\boldsymbol{r}$ and $\boldsymbol{s}$.
The spherical harmonic coefficients $(f*g)_{l,m}$ of $f*g$ satisfies
\begin{equation}
\small{
(f * g)_{l,m} = \left\{ \begin{array}{cc}
\sqrt{\frac{4\pi}{2l+1}} f_{l,m} g^*_{l,0} & l = 0,...,L-1,\ m = -l, ..., l\\
0 & \textrm{otherwise}
\end{array}
\right. }
\label{eq:sphereconv}
\end{equation}
We utilize this observation to design angular responses that are azimuthally-symmetric.
We choose our pixel orientations to measure $f*g$ according to the sampling scheme with fast exact spherical harmonics transform \cite{mcewen2011novel}; this allows us to recover $f_{l,m}$ up to bandlimit $L$.

%% file: orbcam.tex
In this section, we discuss a simple instance of imaging on a curved surface, namely imaging on a sphere. 
We refer to the resulting imager design and its implementation as \textit{OrbCam.}
\subsection{Forward model}
We now derive the measurement model for our imaging system.
The imaging system is modeled as a collection of $N$ photodiodes; all of them the same real-valued angular response function on the sphere $g(\br)$, where $\br = (\theta, \phi)$ in Euler angles.
We limit to the discussion of azimuthally-invariant $g(\cdot)$, \ie $g(\theta, \phi) = g(\theta)$, for tractable analysis.
We assume the scene is a real-valued function on a sphere; $x(\br)$ representing radiance of light from direction $\br$.
This assumption is valid for scene very far away compared to the radius of the spherical sensor. 
%
We also assume the orientations of the pixels $\{(\theta[i], \phi[i]),  i=1, \ldots, N\}$ are known.

Under these assumptions, the measurement made by the $i$-th pixel is the convolution of the scene and the angular response function, $g(\boldsymbol{r})$, evaluated at  $(\theta[i], \phi[i])$, \ie,
\begin{equation}
	y[i] = (x \ast g)(\theta[i], \phi[i]),
	\label{eq:forward}
\end{equation}
where $n[i]$ denotes the measurement noise.

\begin{figure*}[t]
\renewcommand{\arraystretch}{0.7}
\setlength{\tabcolsep}{0pt}

\begin{tabular}{P{0.25\textwidth} P{0.75\textwidth}}
scene & scaling coefficients of simulated angular response\\
\includegraphics[width=0.25\textwidth]{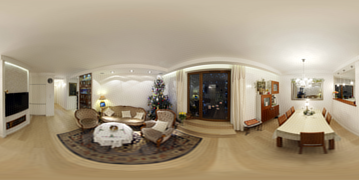} & 
\includegraphics[width=0.7\textwidth]{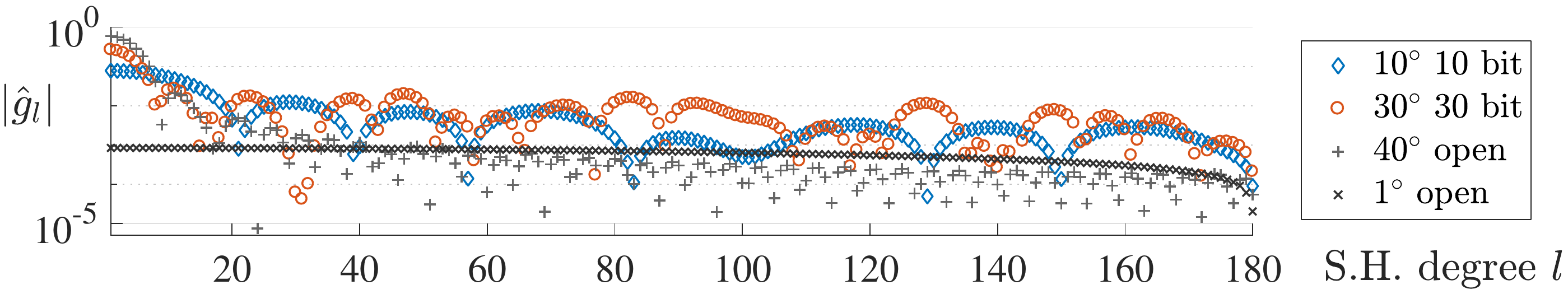} 
\end{tabular}\\

\setlength{\tabcolsep}{-1.5pt}
\begin{tabular}{cccc}
\includegraphics[width=0.26\textwidth]{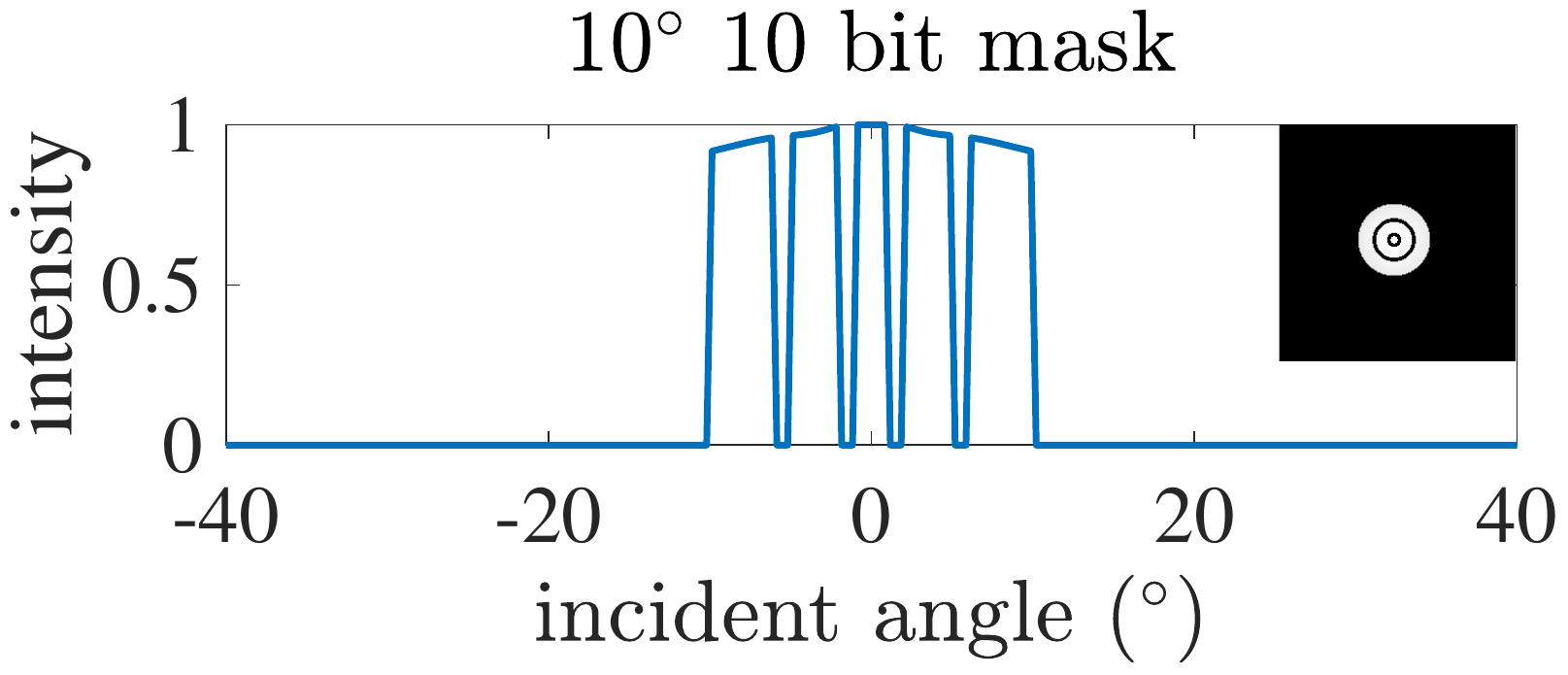} &
\includegraphics[width=0.26\textwidth]{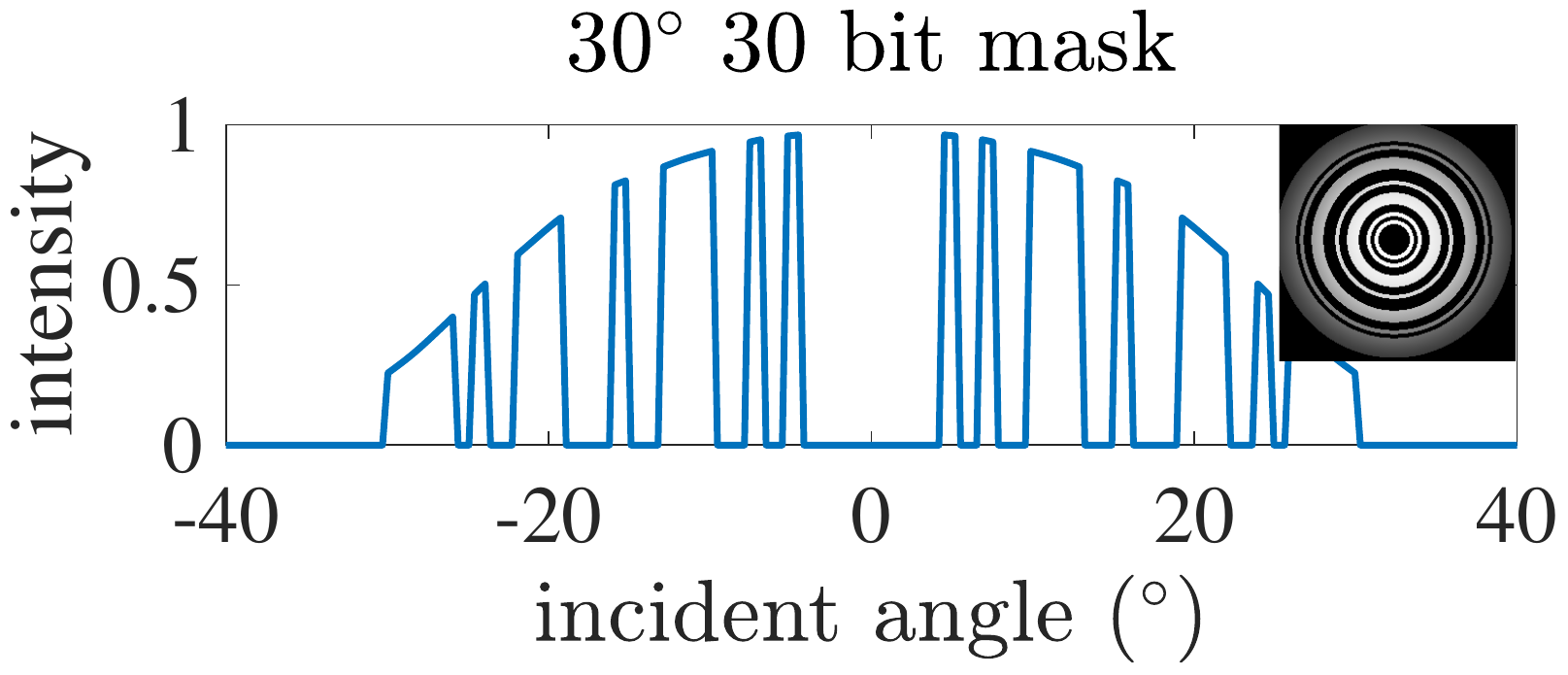} &
\includegraphics[width=0.26\textwidth]{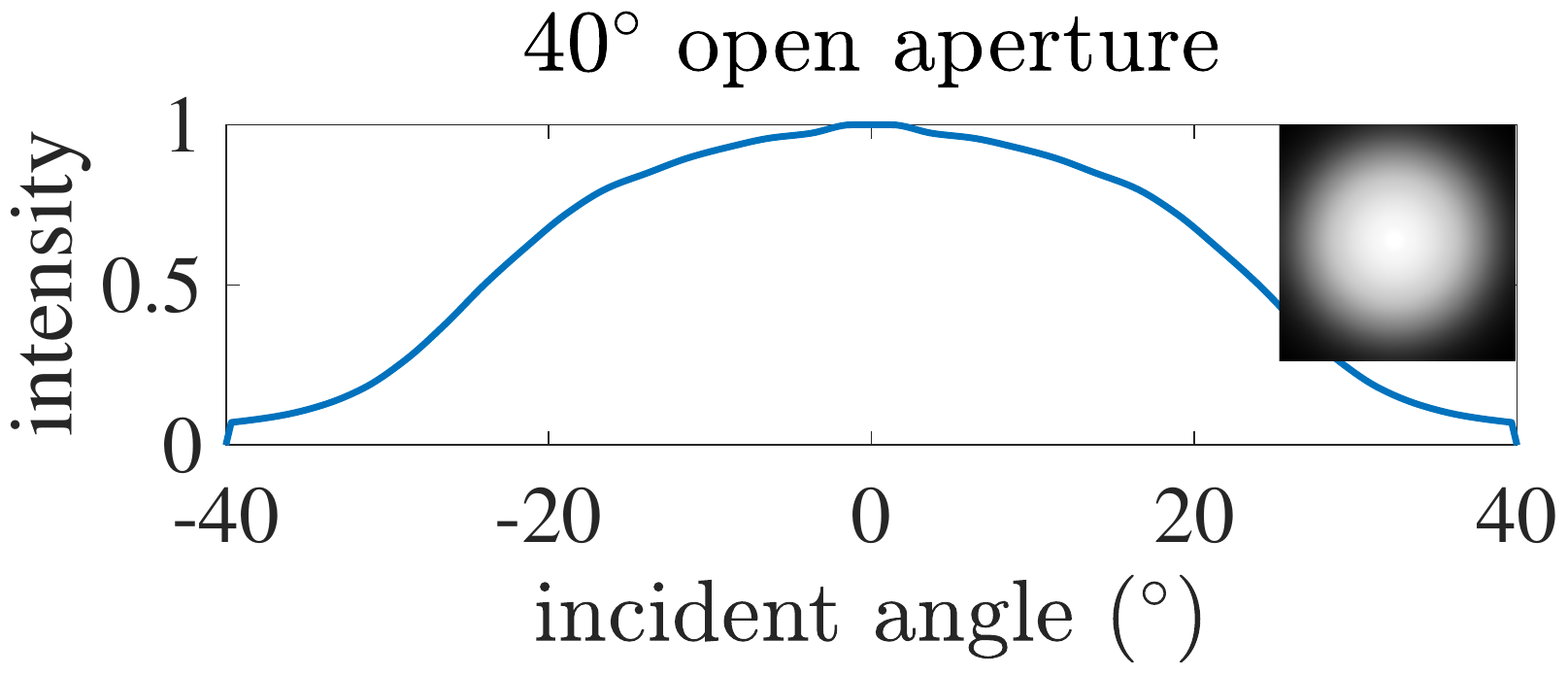} &
\includegraphics[width=0.26\textwidth]{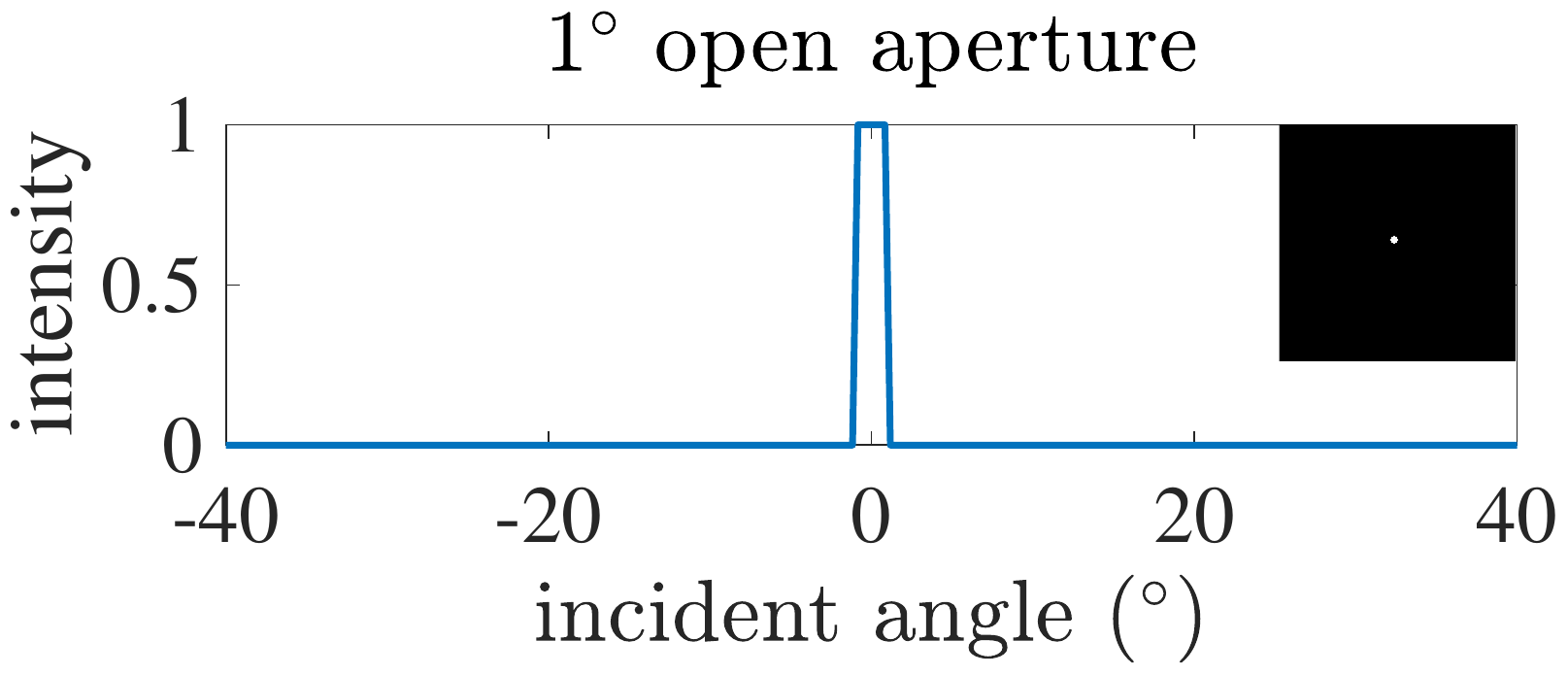}
 \\
\includegraphics[width=0.25\textwidth]{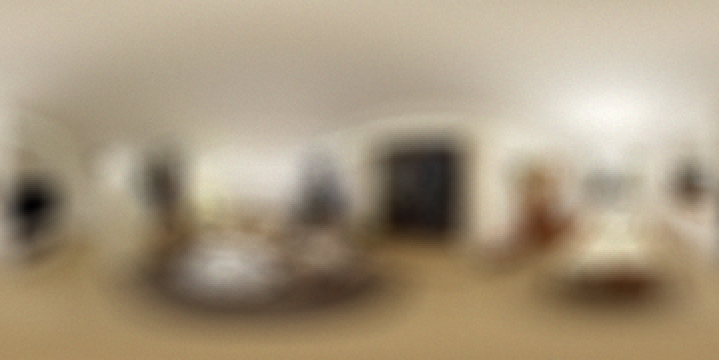} & 
\includegraphics[width=0.25\textwidth]{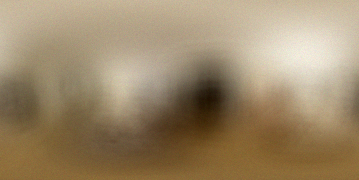} & 
\includegraphics[width=0.25\textwidth]{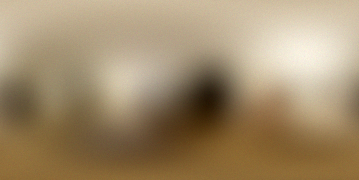} & 
\includegraphics[width=0.25\textwidth]{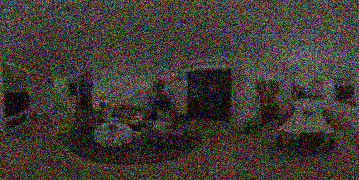} \\
\includegraphics[width=0.25\textwidth]{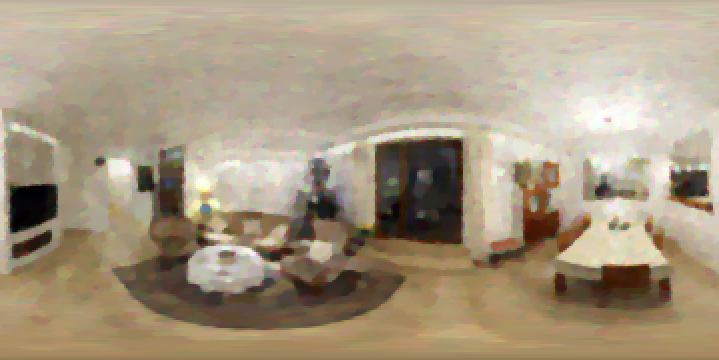} &
\includegraphics[width=0.25\textwidth]{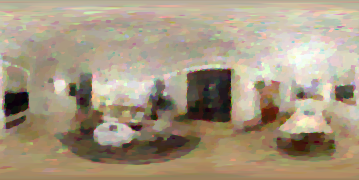} &
\includegraphics[width=0.25\textwidth]{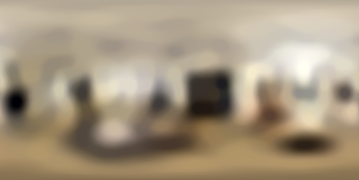} &
\includegraphics[width=0.25\textwidth]{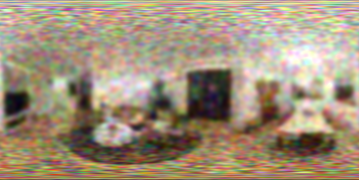} 
\end{tabular}
\caption{
Simulations for comparing effectiveness of different angular responses.
Top left: ground truth scene.
Top right: Norm of scaling coefficients $\hat{g}_l$ of different angular responses. Minimizing expected reconstruction error corresponds to maximizing area under the curve.
Second row: plot of different angular responses; the binary amplitude mask used to produce the angular response is shown in inset.
Third row: measurements obtained with corresponding angular response. Photon noise and readout noise is simulated for a sensor of resolution 180 $\times$ 359 with 32761 $e^-$ saturation capacity and dynamic range of 73.07dB, with scene brightness of 40\%, where 100\% brightness saturates the full well capacity of the sensor with 40~\deg\ open aperture. Intensity is scaled to show noise details.
Last row: images recovered with isotropic total variation prior. Each color channel is individually reconstructed.
}
\label{fig:mask}
\end{figure*}

\subsection{Design of the Pixel Angular Response}
For open apertures, there is a trade off between light throughput and the conditioning of the matrix. 
Large apertures have high light throughput and produces measurements robust to measurement noise, but creates ill-conditioned systems that cannot be inverted.
Small apertures, \ie, pinholes, creates well-conditioned systems but have very low light throughput, and suffers from measurement noise.

We next analyze the performance of imaging with a spherical sensor for a given pixel angular response function $g(\cdot)$ and, subsequently, optimize it for well-conditioned recovery.

Spherical harmonic coefficients of ${y[i]}$ and ${n[i]}$, $y_{l,m}$, $n_{l,m}$, can be computed by spherical harmonics transform given by the sampling scheme.
With the constraint of azimuth symmetric angular response and bandlimit assumptions, the forward imaging process in \eqref{eq:forward} can be rewritten in spherical harmonic domain using \eqref{eq:flm}, \eqref{eq:f_bandlimit}, \eqref{eq:g_az}, and \eqref{eq:sphereconv} into 
\begin{equation}
	y_{l,m} = \widehat{g}_l f_{l,m} + n_{l, m},
	\label{eq:forward_freq}
\end{equation}
where $\widehat{g}_l = \sqrt{\frac{4\pi}{2l+1}g_{l,0}^*}$ for $l=0, \dots, L-1$, and $m=-l, \dots, l$.
An estimate of $f_{l,m}$ by
\begin{equation}
	\widehat{f}_{l,m} = y_{l,m} / \widehat{g},
	\label{eq:reconstruct}
\end{equation}
results in overall error,
\begin{equation}
	\text{error} = \sum_{l, m} \vert \widehat{f}_{l,m} - f_{l,m}\vert ^2 
	= \sum_{l, m} \vert \widehat{g}_l^{-1} \vert^2 \vert n_{l,m} \vert^2
\end{equation}

\paragraph{Metric for optimizing pixel angular response.} Assuming additive white Gaussian sensor noise, minimizing the expected value of error is equivalent to maximizing
\begin{equation}
	\text{mask robustness}(g) = \left(\sum_{l=0}^{L-1} \vert \widehat{g}_l^{-1} \vert ^2 \right)^{-1}
	\label{eq:robustness}
\end{equation}
for recovering scene within a bandlimit of $L$ levels. 
Depending on the method used to modify angular response function, we can search the feasible pixel angular response under fabrication constraints, and select the mask that maximizes the mask robustness function.

\paragraph{Searching for the optimal binary amplitude masks.}
Placing an amplitude mask that blocks light on top of each pixel is a easy way of modifying the angular response function.
Binary masks can be laser printed on thin films with feature size around 10{\textmu m} and placed very close to the sensor.
While amplitude masks can only reduce the angular response from the cosine shaped angular response of most pixel wells (right column in Fig.~\ref{fig:angresp}), we found it improves the conditioning of the measurement matrix while increasing light throughput, compared to a narrow open aperture.

We search for the optimal mask by calculating the robustness of its resulting angular response, using Eq.~\eqref{eq:robustness}, from binary masks of given length with a given angle of view.
For the results in this paper, we optimized masks for two angle of view: for a 10 degree half-aperture opening, we exhaustively searched all 10-bit binary codes; for the 30 degree half-aperture opening, we used stochastic gradient descent with random initialization to search for 30-bit binary codes.
The resulting angular response functions are plotted in Fig.~\ref{fig:mask}.

\paragraph{Light efficiency of optimal binary amplitude masks.}
Total light throughput can be calculated by integrating angular response function $g(\cdot)$ on the sphere.

\begin{equation}\label{eq:light_throughput}
	\begin{split}
		\text{light throughput}(g) &= 
		\int_{\boldsymbol{r} \in \mathbb{S}^2} g(\boldsymbol{r}) d\boldsymbol{r}\\
		&= \sum_{l,m} g_{l,m} \int_{\boldsymbol{r} \in \mathbb{S}^2} Y_{l,m}(\boldsymbol{r}) d\boldsymbol{r}\\
		&= \sqrt{4\pi} g_{0,0}
	\end{split}
\end{equation}
The unmodified angular response function on the sensor we did experiments with results in light throughput of 0.59 steradian from our measurements.
The 10\deg\ 10-bit mask shown in Fig.~\ref{fig:mask} has light throughput of 0.09 steradian, i.e. 15.2\% of unmodified light throughput.
The 30\deg\ 30-bit mask has light throughput of 0.20 steradian, i.e. 33.9\% of unmodified light throughput.

\subsection{Reconstruction}

Most deconvolution algorithms can be extended to recover the original scene convolved with the designed angular response function.
In the absense of noise, the scene $f$ can be recovered from its spherical harmonics coefficients $\widehat{f}_{l, m}$ estimated by Eq.~\eqref{eq:reconstruct}.
When noise is present and we have measurement on all sampling orientations, we can use similarly fast filter approaches such as Wiener filtering.
When we undersample the measurements, as in the case when we only image on parts of a sphere, we can recover the scene robustly with image priors by
\begin{equation}
	\min_{\bff\in\mathcal{R}^+} \text{prior}(\bff) \text{ such that } \Vert \bPhi \bff - \by \Vert^2 < \epsilon,
\end{equation}
where $\bPhi$ is the measurement matrix, and $\epsilon$ is a upper bound on sensor noise.
Iterative approaches are commonly used for solving this optimization problem, and our formulation of the forward process is isotropic convolution on the sphere makes it possible to adopt such iterative approaches.
The reconstruction results in this paper are solved using isotropic total variation on the sphere \cite{mcewen:css2} as prior, and we implmented MFISTA as described in \cite{beck2009fast} to optimize the objective,
\begin{equation}
	\argmin_{ \bff \in \mathcal{R}^+} \Vert \bPhi \bff - \by \Vert^2 + \lambda \text{TV}_{\text{isotropic}}(\bff).
\end{equation}

\subsection{Performance of different angular response functions}
\begin{figure}[!ttt]
	\begin{center}
		\includegraphics[width=0.35\textwidth]{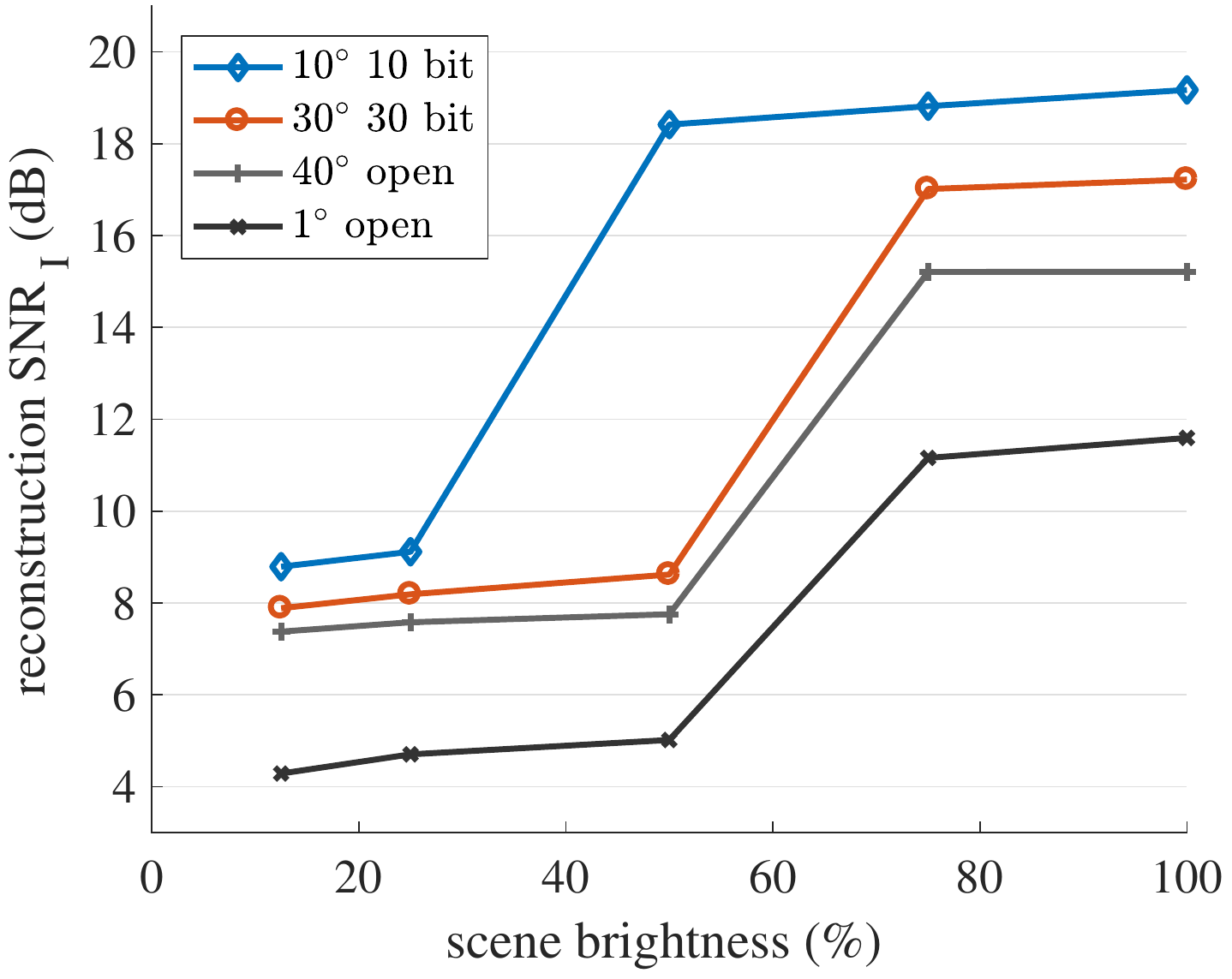}
	\end{center}
	\caption{Average reconstruction $SNR_{I}$ for scene of different brightness over three runs. Measurements simulated with photon noise and readout noise of a sensor with full well capacity 32761 $e^-$ and dynamic range of 73.07dB. Scene at 100\% brightness saturates pixel's full well capacity at the brightest pixel with $40^\circ$ open aperture.}
	\label{fig:rnsr_noise}
	\vspace{-2mm}
\end{figure}

We demonstrate the robustness of our 10-bit and 30-bit binary masks by simulating reconstruction from noisy measurements obtained with this mask, compared against 40\degree\ large aperture and a small 1\degree\ aperture, using open angular response measured from a photodiode.
Specifications of sensors are used to generate noisy measurements:
Photon noise is computed from full well capacity, and the readout noise is computed from dynamic range.
Some simulated measurements, along with their reconstructions and the ground truth scene are shown in Fig.~\ref{fig:mask}.
Quantitatively, the reconstruction signal to noise ratio compared to the original scene is plotted for scene of different brightness in Fig.~\ref{fig:rnsr_noise}.

The reconstructed signal to noise ratio is evaluated using
\begin{equation}
	\text{SNR}_I = 10 \log\left(\frac{\bff^T Q \bff}{\left(\widehat{\bff}-\bff\right)^T Q \left(\widehat{\bff}- \bff\right)}\right),
\end{equation}
where $Q$ contains the quadrature weights given by the sampling scheme \cite{mcewen:css2} to account for sampling density on different parts of the sphere.
We observe that the optimized mask outperforms small apertures in reconstruction SNR.

\begin{figure*}[t]
\hspace{-3mm}
\setlength{\tabcolsep}{1pt}
\begin{tabular}{cccc}
\includegraphics[width=0.25\textwidth]{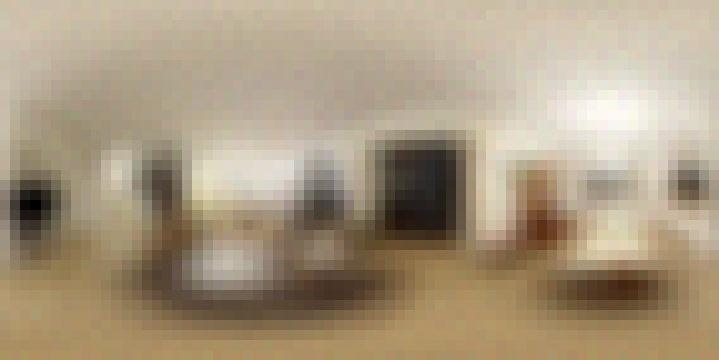} & 
\includegraphics[width=0.25\textwidth]{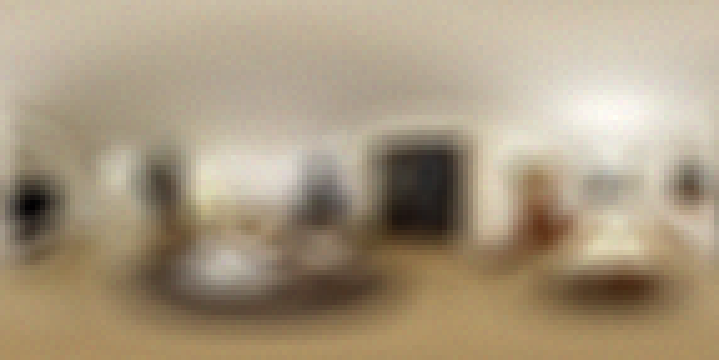} & 
\includegraphics[width=0.25\textwidth]{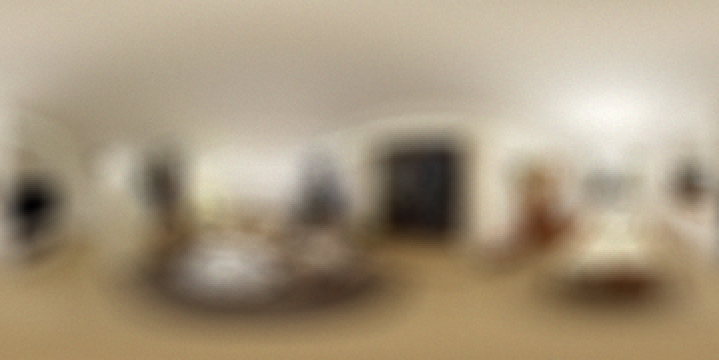} & 
\includegraphics[width=0.25\textwidth]{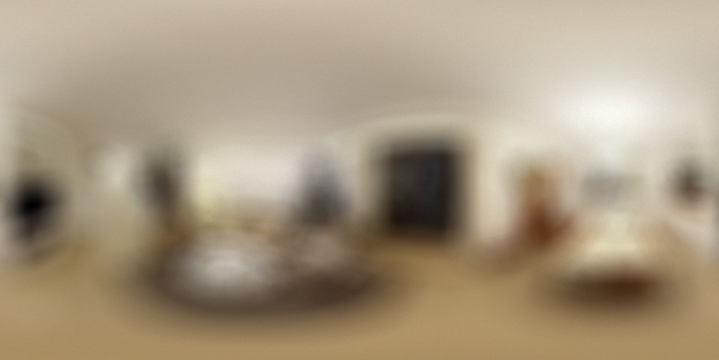} \\
\includegraphics[width=0.25\textwidth]{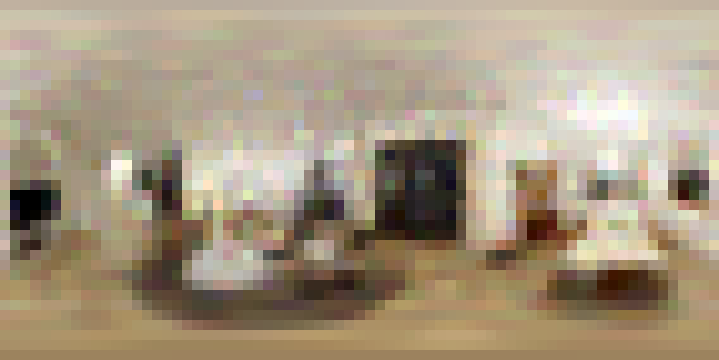} &
\includegraphics[width=0.25\textwidth]{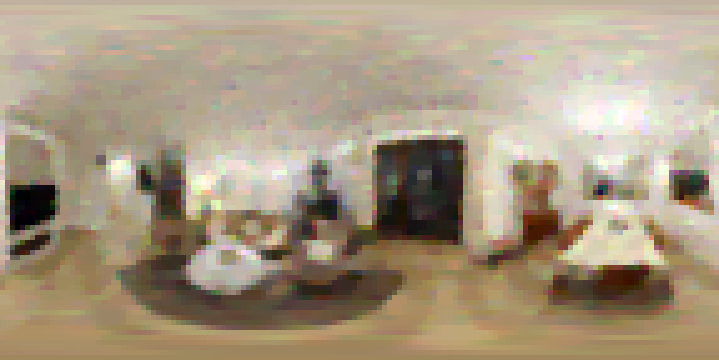} &
\includegraphics[width=0.25\textwidth]{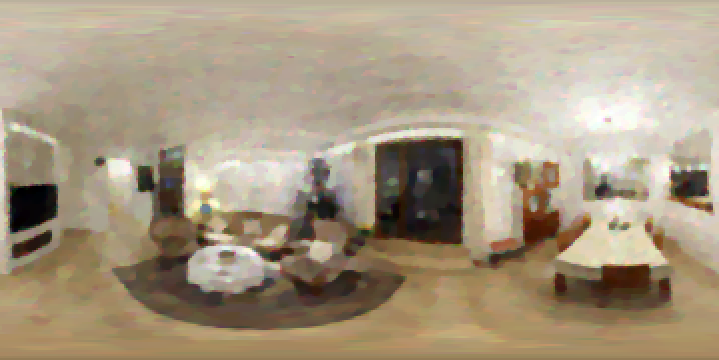} &
\includegraphics[width=0.25\textwidth]{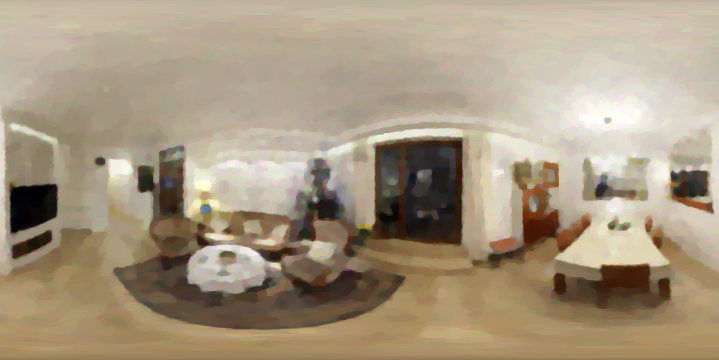} \\
10\deg\ (L=36) &
5\deg\ (L=72) &
2\deg\ (L=180) &
1\deg\ (L=360)
\end{tabular}
\caption{
Simulations for capturing images at different resolution.
Top row: measurements obtained with 10\deg 10-bit mask shaped angular response on a sensor with resolution L$\times$(2L-1), and the same noise statistics and scene brightness as given in Fig.~\ref{fig:mask}.
Bottom row: reconstructed images.
}
\label{fig:resolution}
\end{figure*}

\begin{figure*}[t]
\hspace{-3mm}
\setlength{\tabcolsep}{1pt}
\begin{tabular}{cccc}
\includegraphics[width=0.25\textwidth]{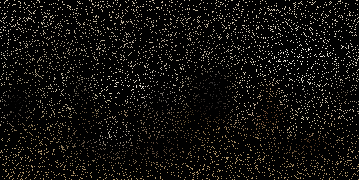} & 
\includegraphics[width=0.25\textwidth]{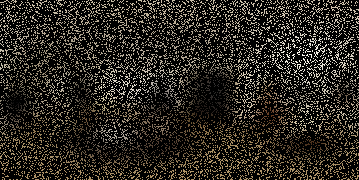} & 
\includegraphics[width=0.25\textwidth]{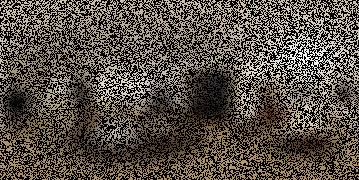} & 
\includegraphics[width=0.25\textwidth]{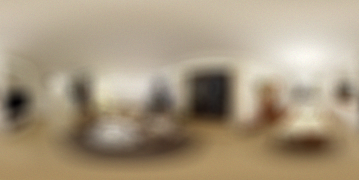} \\
\includegraphics[width=0.25\textwidth]{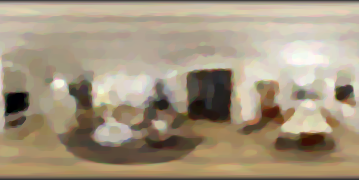} &
\includegraphics[width=0.25\textwidth]{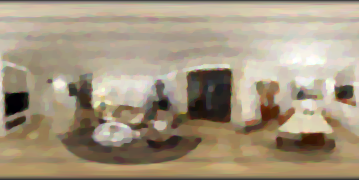} &
\includegraphics[width=0.25\textwidth]{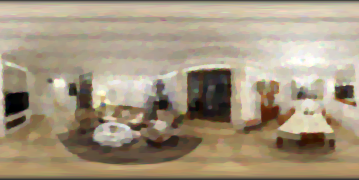} &\includegraphics[width=0.25\textwidth]{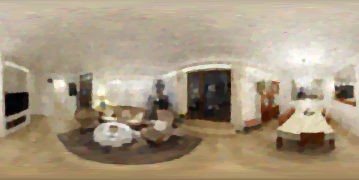} \\
10\% &
25\% &
50\% &
100\%
\end{tabular}
\caption{
Simulations for undersampled images.
Top row: measurements obtained with 10\deg\ 10-bit mask shaped angular response on a sensor with resolution 180$\times$359, and the same noise statistics and scene brightness as given in Fig.~\ref{fig:mask}.
Bottom row: reconstructed images.
}
\label{fig:undersample}
\end{figure*}

%% file: analysis.tex
We conducted experiments to demonstrate that the proposed imager can be used to image planer scenes, complex scenes, and scenes with a large large angle of view.

\begin{figure}[!ttt]
\begin{center}
\includegraphics[trim={2cm 0cm 2cm 0cm}, clip=true, width=0.45\textwidth]{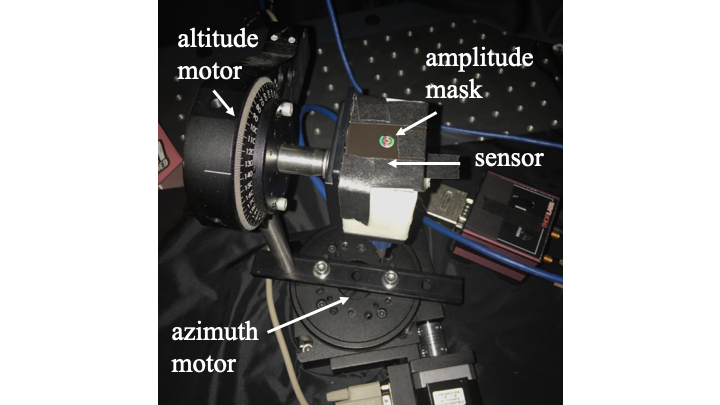}
\end{center}
   \caption{Setup of our prototype experiments.
   A planar sensor (Sony IMX 174) is mounted on an side rotation motor, which is mounted on a rotation stage to produce measurements from a spherical sensor.
   An amplitude mask is laser printed on thin film and fixed on top of the sensor.
   The center region of the sensor is used to produce measurement of one pixel at a given orientation.
   }
\label{fig:prototype}
\vspace{-2mm}
\end{figure}

\paragraph{Prototype.}  To emulate the working of a spherical sensor, we built a two degree of freedom  stage using two rotation motors and mounted a planar sensor (Sony IMX 174) on top.
The rotation stages allowed the sensor to be oriented, repeatably, over the full 360\degree\ angle of view at a precision of $0.5$\degree.
Amplitude masks are laser printed and affixed to 5mm away from the sensor.
Minimum ring widths on film mask is $\sim$ 90 {\textmu m}.
The image sensor was used as a photodiode by grouping together the central $4 \times 4$ pixels --- this provides an effective sensing area of $23.5 \times 23.5\ \text{\textmu m}^2$.
Given that we had a color sensor, half of these pixels contributed to the green channel and a quarter each toward red and blue channels.
The setup is shown in Fig.~\ref{fig:prototype}.

\begin{figure}[t]
\begin{center}
\setlength{\tabcolsep}{-1.5pt}
\begin{tabular}{cc}
\includegraphics[width=0.53 \linewidth]{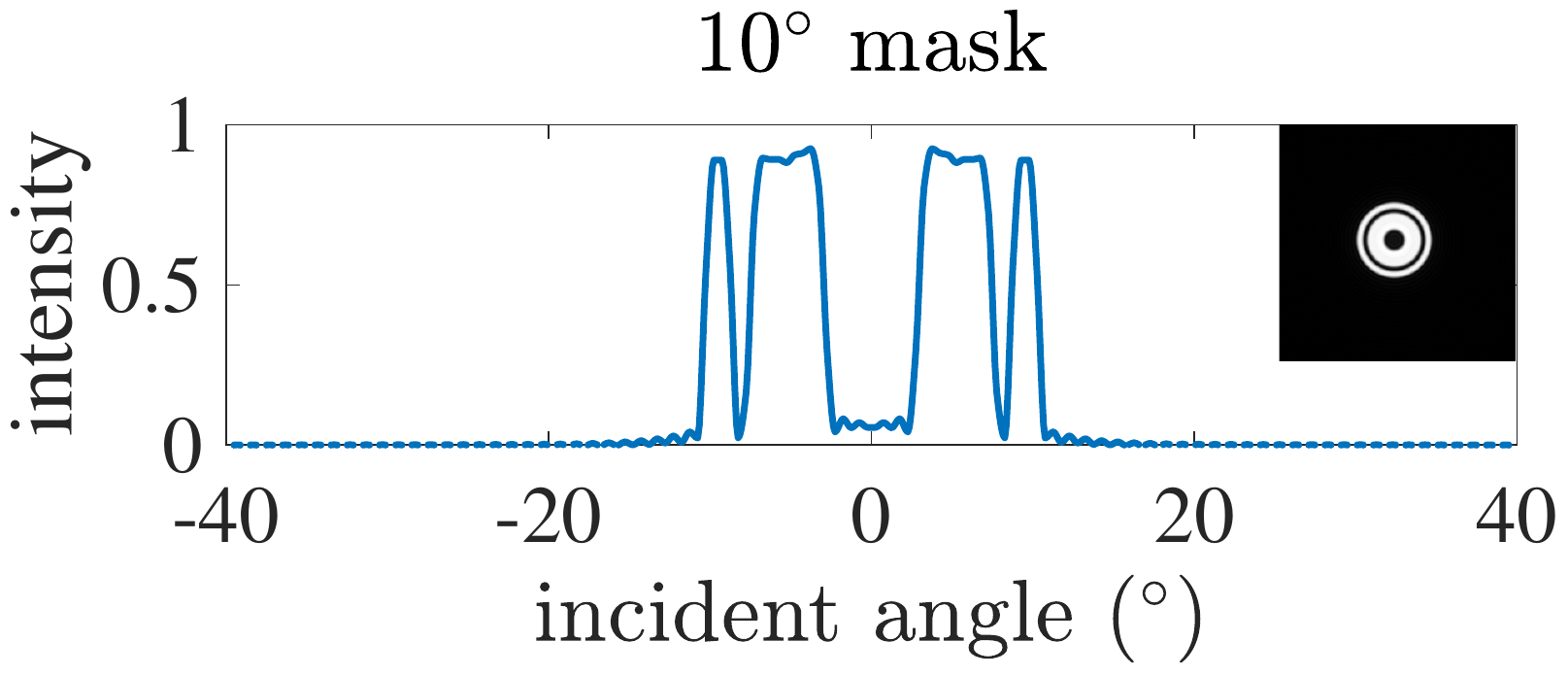}&
\includegraphics[width=0.53 \linewidth]{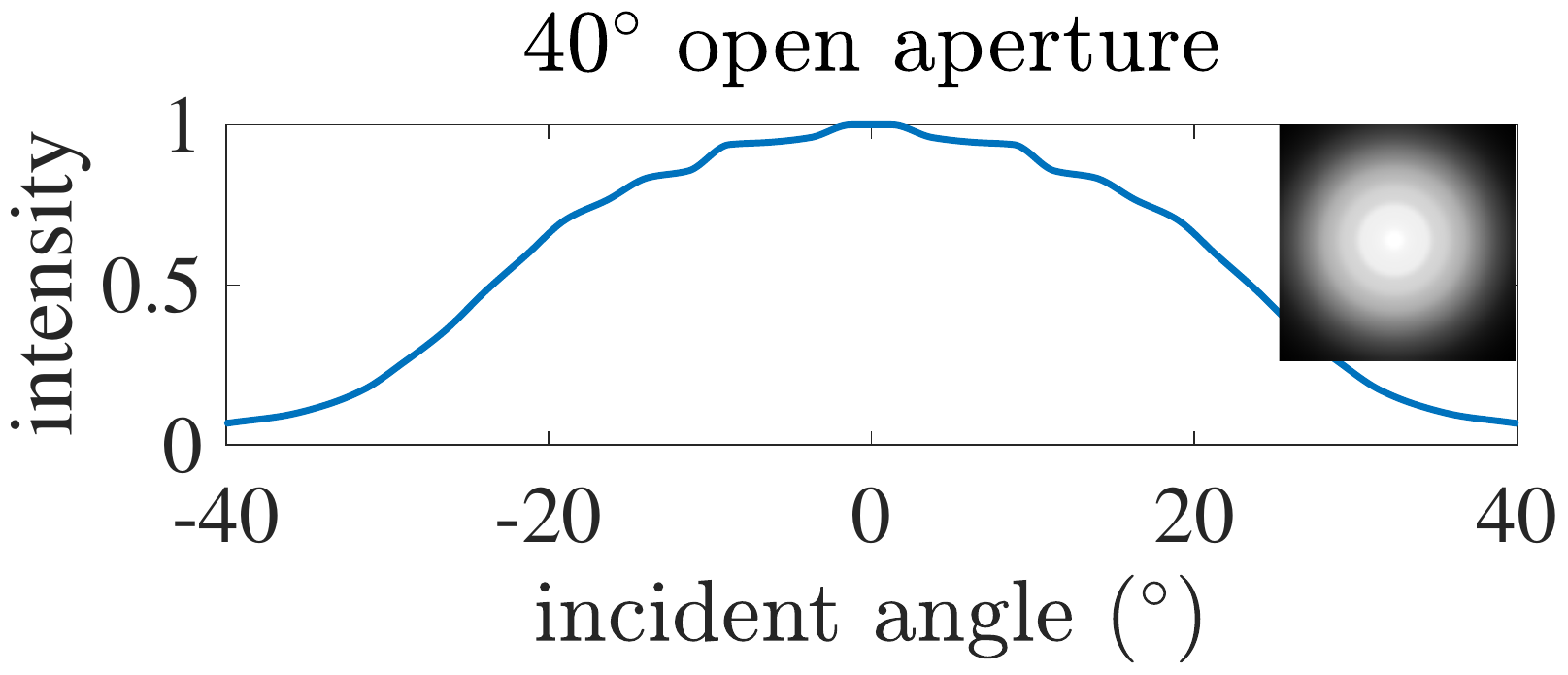} 
\end{tabular}
\caption{
Measured angular response, before and after being modified by binary amplitude mask. Insets show the captured pixel angular response over 80\deg $\times$ 80\deg. The angular response functions are estimated from measurements obtained by rotating the sensor facing a small LED light in a black box. Though unmodeled effects such as diffraction are present, the measured angular responses are still robust to noise.}
\label{fig:angresp}
\vspace{-5mm}
\end{center}
\end{figure}

\paragraph{Calibration.} 
Angular response of a open pixel was measured by rotating a sensor observing a small LED light source and projecting onto closest azimuthally-symmetric function for each color channel.
Fig.~\ref{fig:angresp} show the calibrated angular responses for the a $10$-bit mask pattern, as well as 40\deg\ aperture.
Note the $10$-bit mask used in real experiments differs from the optimized $10$-bit mask shown in the rest of the paper.
We observe diffraction effects and other secondary effects not modeled in our search for optimal mask.
Yet once we calibrated the masks, the scaling coefficients of the measured angular response indicates the angular response we managed to produce are still robust.
Our architecture allows angular response shaping by reflection, diffraction, or refraction.

\paragraph{Reconstructions.}
We scan a range of scenes with different angle of views and sampling resolutions.
Due to moving components on the experiment prototype,  we were able to reconstruct partial scenes on the sphere with the largest being one-third of the whole sphere (see Fig.~\ref{fig:teaser}).

\begin{figure*}[t]
\begin{center}
\includegraphics[page=1, trim={0cm 6.5cm 0cm 0cm}, clip=true, width=0.9\textwidth]{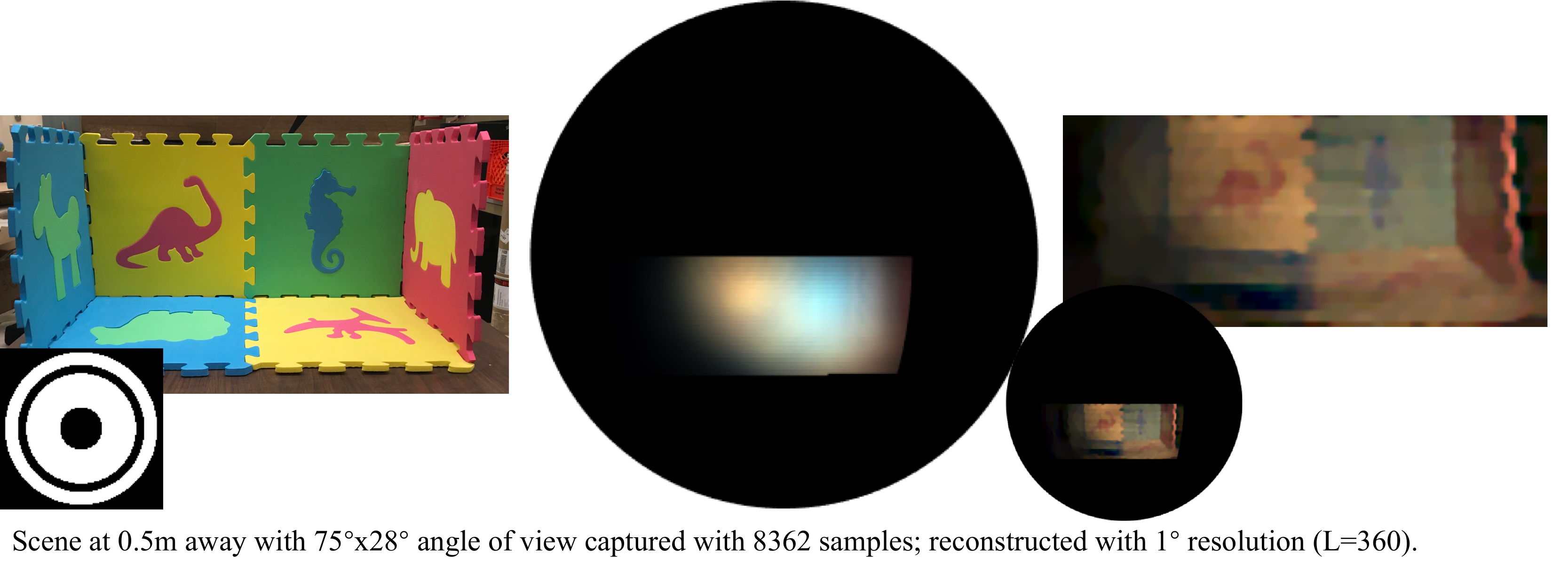}
  \includegraphics[page=3, trim={0cm 6.5cm 0cm 0cm}, clip=true, width=0.9\textwidth]{figures/figure_gen.pdf}
 \includegraphics[page=2, trim={0cm 6cm 0cm 0cm}, clip=true, width=0.9\textwidth]{figures/figure_gen.pdf}
  \end{center}
   \caption{Prototype experiment results. Left: scene setup; inset shows the mask pattern used to modify angular response of the pixel. Middle: captured measurements. Right: reconstruction of the scene.}
\label{fig:exps1}
\vspace{-2mm}
\end{figure*}

In Fig.~\ref{fig:exps1},  we present scenes with different angle of view, depth, and structure.
For one of the scene, we also reconstruct from measurements spanning a random subset of orientations; given that the measurements from neighboring orientations are highly redundant, this does not cause a significant drop in performance.

%% file: discuss.tex
In this paper, we present the design of a lensless imager that consists of a spherical sensor where-in each pixel has an identical but optimized angular response function.
This design enables lensless imaging with a new capability --- namely, imaging on a spherical surface.
We propose a metric to evaluate the optimality of pixel angular response functions for robust recovery of image in the presence of noise, and demonstrated the validity of our design with simulation and real experiments.

\subsection{Limitations}
Our current implementation has two key limitations.
First, we placed an amplitude mask in front of the sensor with a standoff distance of about 5mm.
Clearly, this design does not scale when we have high resolution spherical sensor arrays.
For such arrays, the mask needs to be embedded on top of each pixel and this requires the design of a diffraction element that can provide the desired angular response.
A second limitation, also stemming from the use of amplitude mask, is the loss of light due to the mask itself.
This could potentially be addressed via phase masks that redistribute light instead of blocking it.
\subsection{Extension to curved and flexible surfaces}
The proposed design can still image when it takes the shape of other curved surfaces, as long as the diversity in pixel orientations is maintained.
We simulate the situation where the designed spherical sensor is deformed to other curved surfaces in Fig.~\ref{fig:flexible}.
\input{figure_flexible.tex}

%% file: figure_flexible.tex
\begin{figure*}[t]
\setlength{\tabcolsep}{1pt}
\begin{tabular}{ccc}
\includegraphics[width=0.25\textwidth]{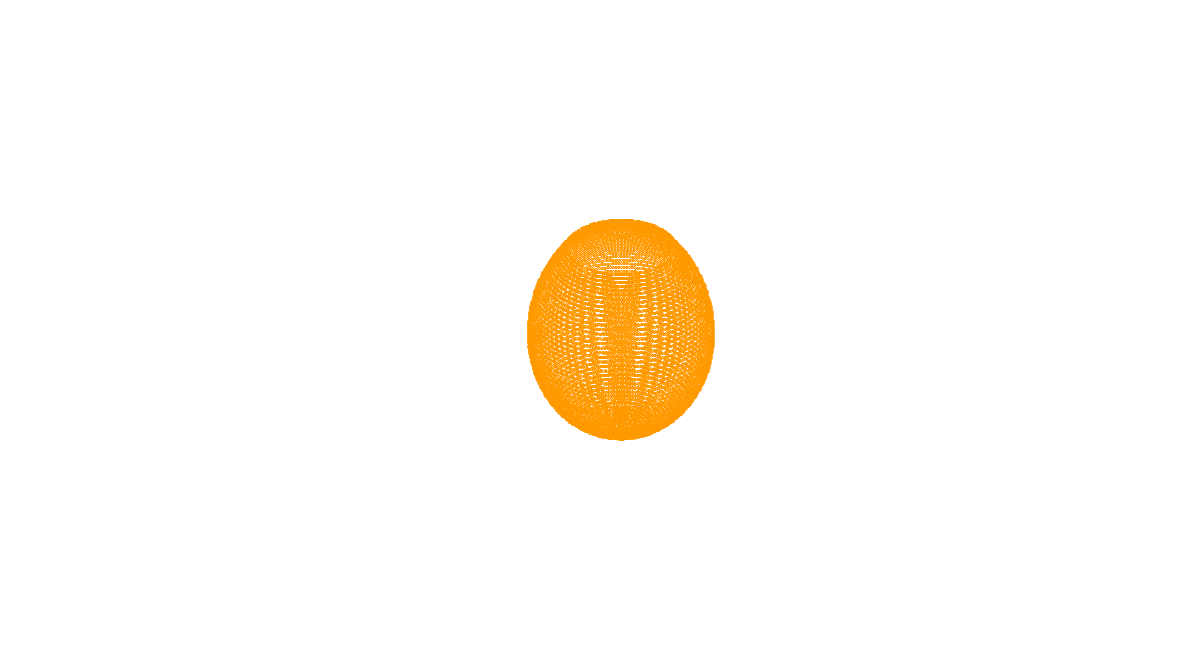} & 
\includegraphics[width=0.35\textwidth]{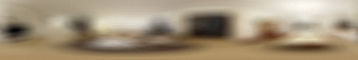} & 
\includegraphics[width=0.35\textwidth]{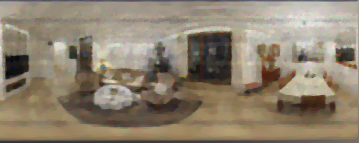} \\
\includegraphics[width=0.25\textwidth]{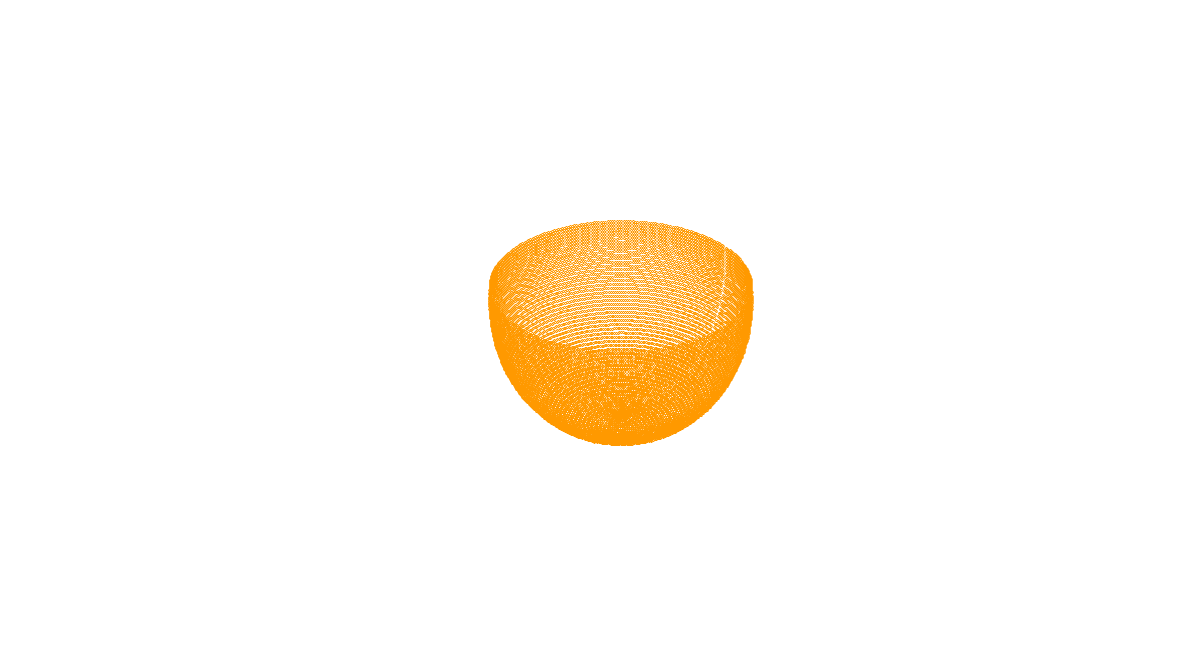} & 
\includegraphics[width=0.35\textwidth]{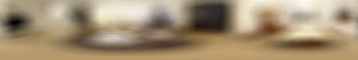} & 
\includegraphics[width=0.35\textwidth]{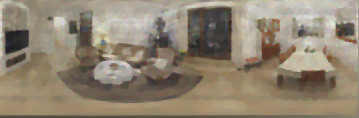} \\
\includegraphics[width=0.25\textwidth]{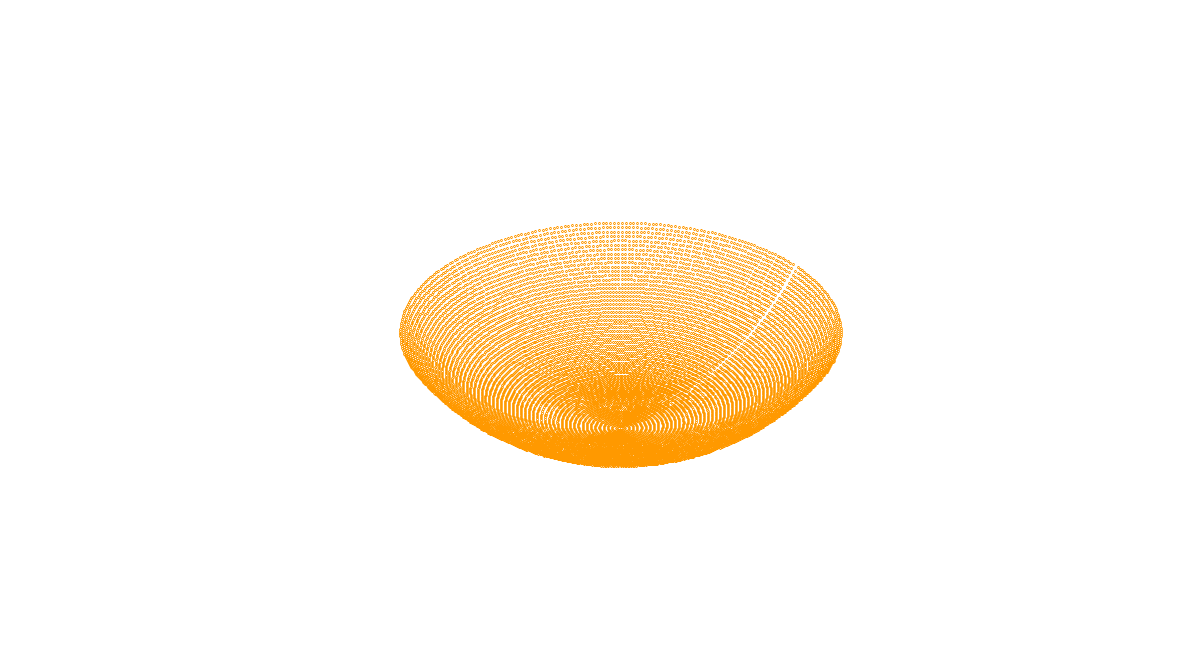} & 
\includegraphics[width=0.35\textwidth]{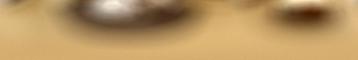} & 
\includegraphics[width=0.35\textwidth]{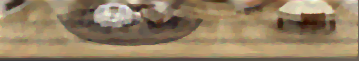} \\
sensor configuration &
measurements &
reconstruction
\end{tabular}
\vspace{2mm}
\caption{
	Simulations for imaging with a few different configurations of a flexible sensor.
	A flexible sensor with proposed angular response function is placed into the scene shown in Fig.~\ref{fig:mask}. 
	The top row shows the spherical sensor with dots representing pixels, each pointing away from the center of the sphere.
	The sensor array is gradually deformed, seeing different parts of the scene.
	The sensor configuration, i.e. the orientation of each pixel, is assumed to be known.
	The proposed imager can stably reconstruct scene under different sensor configurations when there is no self-occlusion.
}
\label{fig:flexible}
\end{figure*}